\documentclass{article}

\usepackage{microtype}
\usepackage{graphicx}
\usepackage{booktabs} 

\usepackage{hyperref}

\usepackage[accepted]{icml2024}

\usepackage{amsmath}
\usepackage{amssymb}
\usepackage{mathtools}
\usepackage{amsthm}
\usepackage{ upgreek }
\usepackage{adjustbox}
\usepackage{tabularx}
\usepackage{makecell}
\usepackage{multirow}
\usepackage{bold-extra}
\usepackage{appendix}

\usepackage{mathrsfs}
\usepackage{subfig}
\usepackage{algorithm}
\usepackage[capitalize,noabbrev]{cleveref}

\theoremstyle{plain}
\newtheorem{theorem}{Theorem}[section]

\theoremstyle{definition}
\newtheorem{definition}[theorem]{Definition}

\theoremstyle{remark}
\newtheorem{remark}[theorem]{Remark}

\usepackage[textsize=tiny]{todonotes}

\icmltitlerunning{NPAT: Null-Space Projected Adversarial Training Towards Zero Deterioration of Generalization}

\begin{document}

\twocolumn[
\icmltitle{NPAT: Null-Space Projected Adversarial Training Towards Zero Deterioration of Generalization}



\icmlsetsymbol{equal}{*}

\begin{icmlauthorlist}
\icmlauthor{Hanyi Hu}{zjlab}
\icmlauthor{Qiao Han}{zjlab}
\icmlauthor{Kui Chen}{zjlab}
\icmlauthor{Yao Yang}{zjlab}
\end{icmlauthorlist}
\begin{center}
\textbf{\{huhy, hanq, chenkui, yangyao\}@zhejianglab.com}\\
\end{center}

\begin{center}
\textbf{Zhejiang Lab}
\end{center}
\icmlaffiliation{zjlab}{Zhejiang Lab, Hangzhou, Zhejiang Province, China}



\vskip 0.3in
]




\begin{abstract}
To mitigate the susceptibility of neural networks to adversarial attacks, adversarial training has emerged as a prevalent and effective defense strategy. Intrinsically, this countermeasure incurs a trade-off, as it sacrifices the model's accuracy in processing normal samples. To reconcile the trade-off, we pioneer the incorporation of null-space projection into adversarial training and propose two innovative \textbf{N}ull-space \textbf{P}rojection based \textbf{A}dversarial \textbf{T}raining(\textbf{NPAT}) algorithms tackling sample generation and gradient optimization, named \textbf{N}ull-space \textbf{P}rojected \textbf{D}ata \textbf{A}ugmentation (\textbf{NPDA}) and \textbf{N}ull-space \textbf{P}rojected \textbf{G}radient \textbf{D}escent (\textbf{NPGD}), to search for an overarching optimal solutions, which enhance robustness with almost zero deterioration in generalization performance. Adversarial samples and perturbations are constrained within the null-space of the decision boundary utilizing a closed-form null-space projector, effectively mitigating threat of attack stemming from unreliable features. Subsequently, we conducted experiments on the CIFAR10 and SVHN datasets and reveal that our methodology can seamlessly combine with adversarial training methods and obtain comparable robustness while keeping generalization close to a high-accuracy model.
\end{abstract}

\section{Introduction}
\label{submission}

Deep learning models are claimed to be universal function approximator\cite{hornik1989multilayer} and have shown promising capability in fitness on different tasks. Contrarily, deep learning models can be vulnerable to human unnoticeable disturbance on input and generate completely unexpected outcome \cite{szegedy2013intriguing}\cite{biggio2018wild}. Adversarial training methods attempt to leverage model vulnerability under these worst-case attacks. The subtlety is the trade-off between the standard error and robustness error, namely, the error on zero perturbed samples and the error on worst-case perturbed samples. The terminology of the trade-off is interchangeable with generalization and robustness in the literature.

Many previous works have explained and provided theoretical analysis on this trade-off problem. There are two main theories in the literature with one claiming the standard training objective is fundamentally different from that of the adversarial task \cite{tsipras2018robustness}\cite{zhang2019theoretically}\cite{fawzi2018analysis} and the other one arguing that the capacity of the classifier is not large enough for improving robustness while keeping accuracy \cite{nakkiran2019adversarial}. However, the sample separation of different classes for MNIST \cite{deng2012mnist}, CIFAR10 \cite{krizhevsky2009learning} and SVHN \cite{netzer2011reading} have been investigated empirically \cite{yang2020closer} that samples are bound to be classifiable perfectly if these attacks were within a $\upvarepsilon$-ball ($l_{\infty}$ perturbation) less than the smallest inter-class separation. Yet, there is no promising method mitigating the trade-off of accuracy and robustness, but mainly controlling the level of trade-off.
\begin{figure}[t]
  
  \includegraphics[width=\columnwidth]{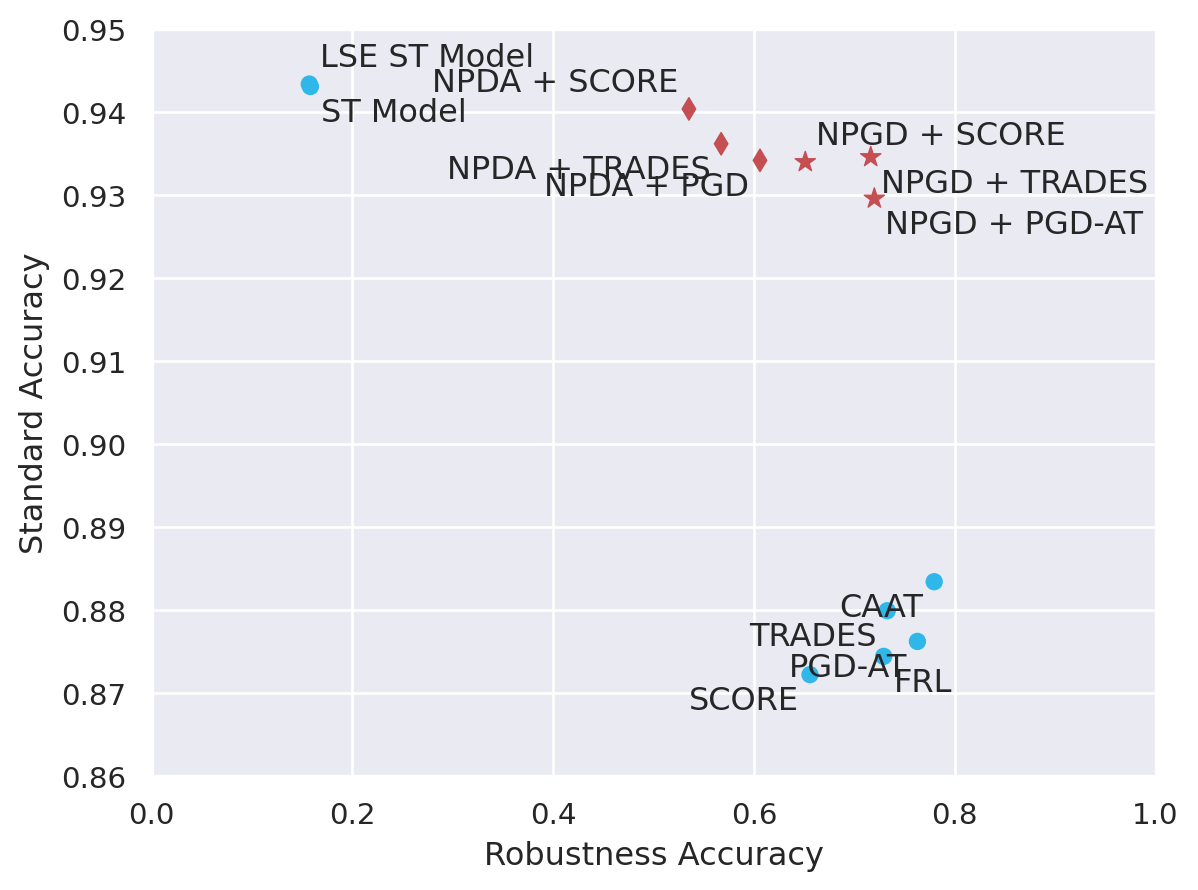}
  \caption{Scatter Plot of Model Standard Accuracy vs. Robustness under Auto-attack on CIFAR10.}
  \label{ScatterPlot}
\end{figure}

\begin{figure*}[t]
\centering
\subfloat[]{
    \label{fig:subfig1}\includegraphics[scale=0.2]{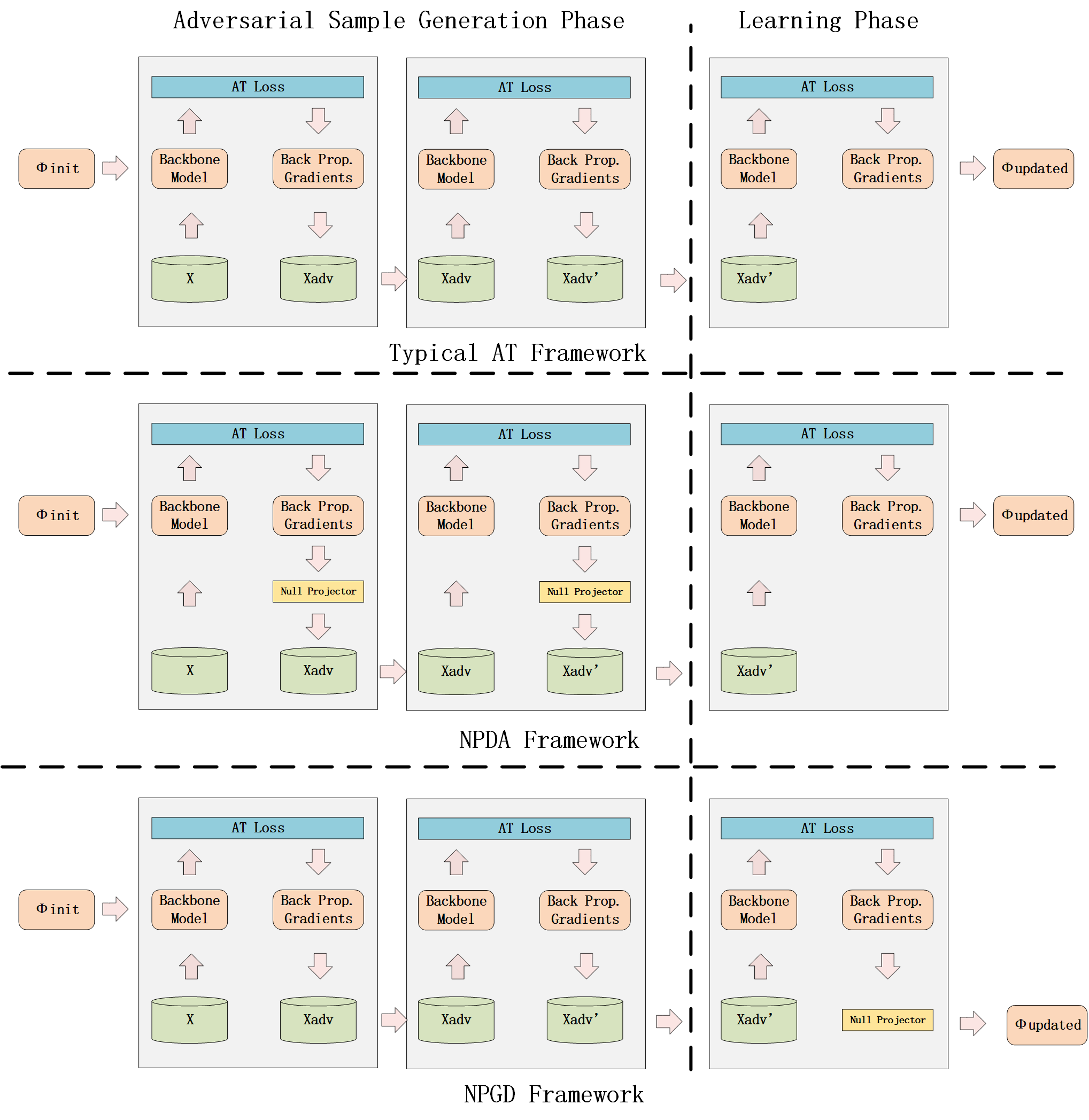}
}
\subfloat[]{
    \label{fig:subfig2}\includegraphics[scale=0.45]{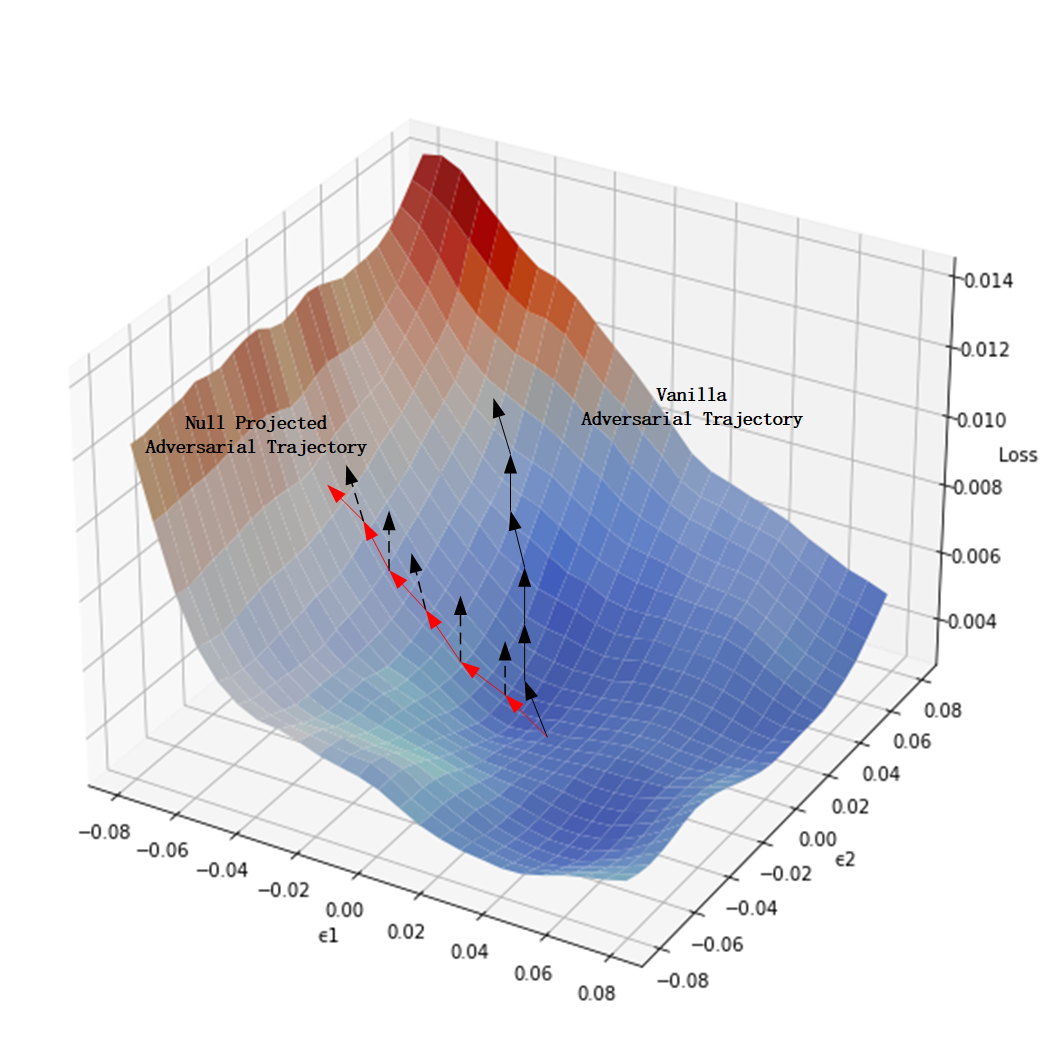}
}
\caption{Illustrations of Null Space Projection-Based Adversarial Training. a) Overall Structure of Adversarial Training Frameworks. b)An Illustration of Multi-step Null-space Projection Sample Generation Process. Black arrows represent the direction of deviation by adversarial training, red arrows represent the direction of null-space projected deviation by adversarial training in NPDA.}
\label{figure2}
\end{figure*}

\begin{figure*}[t]
\centering
    
        \subfloat[Typical AT Top View]{
            \label{fig3:a}\includegraphics[width=0.5\columnwidth]{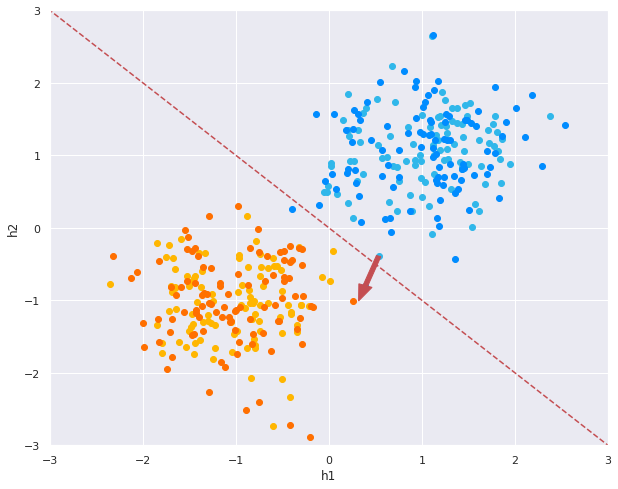}
        }
        \subfloat[Typical AT Front View]{
            \label{fig3:b}\includegraphics[width=0.5\columnwidth]{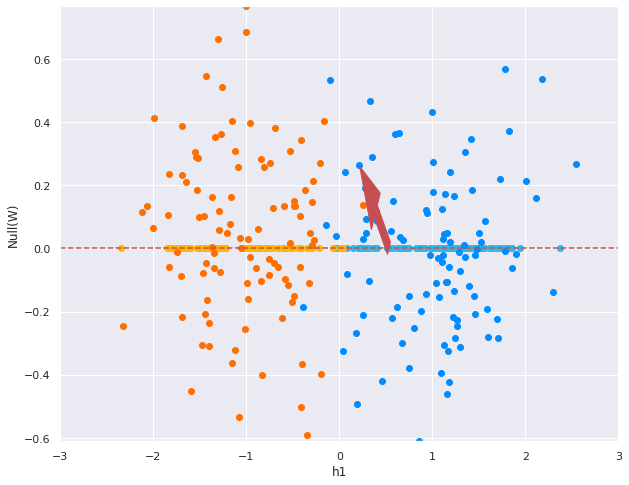}
        }
        \subfloat[NPAT Top View]{
            \label{fig3:c}\includegraphics[width=0.5\columnwidth]{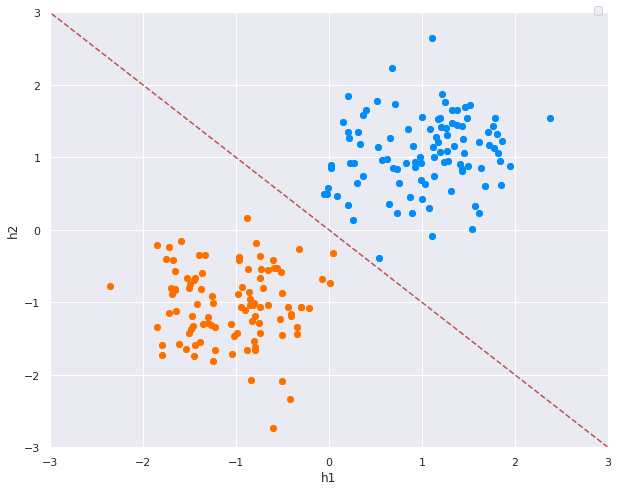}
        }
        \subfloat[NPAT Front View]{
            \label{fig3:d}\includegraphics[width=0.5\columnwidth]{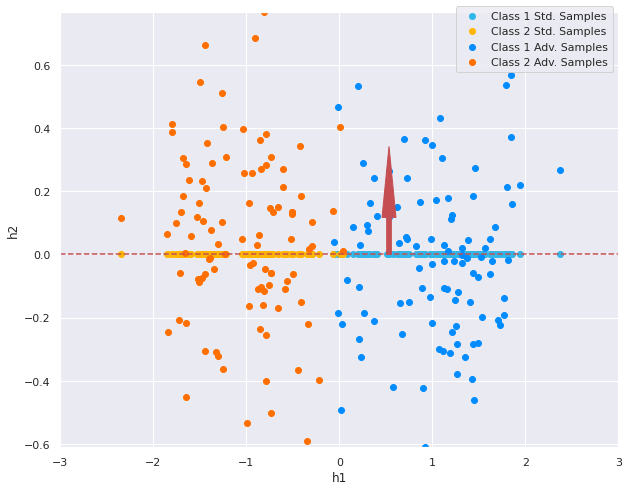}
        }
    
\caption{Distribution Of Toy Sample Representation $y$. Toy distribution of standard and adversarial sample representations from typical adversarial training(Typical AT) \& null projection-based adversarial training(NPAT). Top view is a visualization of two randomly selected dimensions from column space of $W_L$. Front view is a visualization of one randomly selected from column space and one randomly selected dimension from null space of $W_L$. The red arrow denotes the deviation from standard sample to its adversarial peer.}
  \label{ToyDistribution}
\end{figure*}


Attempts on mitigating the standard error and robustness error trade-off are conceptually under three paradigms 1) by introducing extra datasets or data augmentation \cite{carmon2019unlabeled}\cite{najafi2019robustness}\cite{alayrac2019labels} 2) by re-defining boundary loss for robustness \cite{zhang2019theoretically}\cite{pang2022robustness} 3) by weighting loss for different sample due to in-balanced priors, variance, noise-level of each class \cite{xu2021robust}. The null space projection has been deployed by \cite{wang2021training} for increasing model plasticity in continual learning. The objective of utilizing null space projection is to preserve the model ability in the previous task and adapt to another task in the meantime. We propose an estimated null-space projector based adversarial training method mitigating the trade-off without extra data. 
Our contributions are as follows:

\begin{itemize}
    \item We propose two implementations via an estimated null-space projector based on a pre-trained high-accuracy model, which can effectively perform as a seamless add-on to existing adversarial training scope.
    \item Both our null-space projector based methods achieve almost zero deterioration of generalization and boost robustness without extra synthetic dataset and model capacity.
    \item We attempted to manifest these two methods with theoretical analysis and empirical experiments on two open-access datasets CIFAR10 and SVHN to assure the effectiveness of our proposed methods under different settings.
\end{itemize}

The paper is unfolded as follows. We first introduce notations and preliminaries for adversarial training and null-space in section 2. In section 3, we present our adversarial training methods in detail and theoretical analysis between them. The experiment setup and corresponding evaluation are in Section 4 and Section 5. Lastly, we summarize related background works in Section 6. 

\section{Notation and Preliminaries}
\subsection{Standard error, Robust error, Consistent Perturbation} Given an n pair of input $x^{std}\in X^{std} \subseteq R^{n\times d}$ and target $y\in Y \subseteq C^{n\times1}$ dataset $D$, a standard training tend to learn a mapping $f(\cdot; \theta^{std}):X^{std}\rightarrow Y$ with the lowest standard error $\mathcal{L}^{std}$, where $\mathcal{C}$ denotes the target set $\{1,2,…,c\}$.

\begin{equation}
  \begin{aligned}
  \label{standard training loss}
\mathcal{L}^{std}&=-\mathbb{E}_{(x^{std}, y)\sim D}[l(f(x^{std}),y)]
\end{aligned}
  \end{equation}

A typical adversarial training method attempts to optimize the robustness error $\mathcal{L}^{robust}$ with adversarial sample $x^{adv} \in X^{adv} \subseteq R^{n\times d}$ by a consistent perturbation $\mathcal{T}: X^{std} \rightarrow X^{adv}$. Typically, $\mathcal{T}$ often takes imperceptible changes $\delta$ in the original input, such as small affine transformation, contrast changes or small $l_{\infty}$ disturbance derived from input. $\mathcal{T}: x^{adv}=x^{std} + \delta$.\\
The robustness error of adversarial training utilizes adversarial training samples $x^{adv}$ instead and the adversarial loss was initially defined as Eq. \eqref{PGD robost training loss} by \cite{madry2017towards}.

\begin{equation}
  \begin{aligned}
  \label{PGD robost training loss}
\mathcal{L}^{Madry}&=-\mathbb{E}_{(x^{adv}, y)\sim D}[\max_{\delta} l(f(x^{adv}),y)]
\end{aligned}
  \end{equation}

Alternatively, the robustness error can be characterized as a standard classification error term and a boundary error term for robustness as \cite{zhang2019theoretically}.

\begin{equation}
  \begin{aligned}
  \label{TRADES robost training loss}
  \mathcal{L}^{TRADES}&=-[\mathbb{E}_{(x^{std}, y)\sim D}[l(f(x^{std}),y)]\\ &+ \beta \cdot \mathbb{E}_{ (x^{std}, x^{adv})}[Div(f(x^{std}),f(x^{adv}))]]
\end{aligned}
  \end{equation}
where $\beta$ stands for the adversarial coefficient of robustness error balancing the trade-off between two errors and $Div$ is a distance function such as Kullback-Leibler Divergence.

Our goal is to train a model $f(\cdot; \theta^{adv})$ to optimize robustness error $\mathcal{L}^{robust}$, while keeping the standard error $\mathcal{L}^{std}$ close to that of $f(\cdot; \theta^{std})$, a high accurate model trained by standard training configuration according to Eq.\eqref{standard training loss}. Hence, we can define the objective function in the general form as a standard adversarial loss $\mathcal{L}^{robust}$ with a constraint which generate identical output as from a high accurate model $f(\cdot; \theta^{std})$, 
\begin{equation}
  \begin{aligned}
  \label{NSAT robost training loss general form}
\mathcal{\hat{L}}^{robust}&=\min_{\theta^{adv}}\mathcal{L}^{robust}(\cdot) \\
s.t. & \quad f(x^{std};\theta^{adv})= f(x^{std}; \theta^{std})
\end{aligned}
  \end{equation}
The $\theta^{adv}$ stands for the model parameter we try to optimize for robustness, while the $\theta^{std}$ is the parameter trained from a high accuracy model without adversarial setting.

\subsection{Null Space Definition}
\begin{definition}
Given a matrix $W\in R^{d1 \times d2}$, the null space of $W$ is defined as $Null(W)=\{x | Wx=0\}$. \\
\end{definition}

\begin{definition}
Given matrix $W\in R^{d1 \times d2}$ and $r(W) < min\{d1, d2\}$, $\exists$  $P_{Null(W)}$ satisfies that,
\begin{equation}
  \begin{aligned}
     WP_{Null(W)}x=0, \quad for \quad \forall x\in R^{d}  \\
\end{aligned}
  \end{equation}
\end{definition}

If rank of matrix $r(W) < min\{d1, d2\}$, the null space projection matrix $P_{Null(W)}$ exists non-zero closed-form solution. The null space projection matrix is defined as,
\begin{equation}
  \begin{aligned}
  \label{closed form null space projector }
P_{Null(W)} = I - W(W^TW)^{-1}W^{T}
\end{aligned}
  \end{equation}

The computation of $(W^TW)^{-1}$ is costly and $P_{Null(W)}$ is typically solved by Singular Vector Decomposition(SVD). The SVD factorizes a matrix $W=U\Sigma V^T \in R^{m \times n}$, where $U\in R^{m\times m}$ corresponds to orthonormal basis of the column space of $W$, $\Sigma \in R^{m \times n}$ is a pseudo-diagonal matrix. The diagonal elements are the singular values of $W$. $V^T \in R^{n\times n}$ is the orthonormal basis of row space of $W$. The projection of row space of $W$ can be represented as $VV^T$. The projection of null space can be calculated as,

\begin{equation}
  \begin{aligned}
  \label{SVD null space solution}
P_{Null(W)}=I-VV^T
\end{aligned}
  \end{equation}

\subsection{A Closer Look at Model Behavior in Standard Training vs. Advesarial Training} Consider a deep learning model $f_\theta^{dl}$ under standard training, the output $y$ is computed by $L-1$ layer of non-linear transformation denoted as $\varphi(\cdot)$ and a fully-connected transformation $W_L^T$  mapping to the number of classes.

\begin{equation}
  \begin{aligned}
  \label{std model expansion}
h_{L-1}^{std}&= \varphi(x^{std})\\
y^{std}&=W_L^T h_{L-1}^{std}+b
\end{aligned}
  \end{equation}
where $h_{L-1}^{std}$ denotes the output of non-linear transformation.
When the model $f_\theta^{dl}$ is exposed to a consistent imperceptible perturbation $\mathcal{T}$, the output can be represented as,

\begin{equation}
  \begin{aligned}
  \label{adv model expansion}
h_{L-1}^{adv}&= \varphi (x^{adv})\\
        &= \varphi (x^{std}+ \delta)\\
y^{adv} &= W_L^T h_{L-1}^{adv}+b\\
        &= W_L^T (h^{std}+\Delta h^{adv})+b\\
        &= W_L^T h^{std}+W_L^T \Delta h^{adv}+b
\end{aligned}
  \end{equation}

where $\Delta h^{adv}$ is the change in the penultimate stemming from the adversarial perturbation.\\

\section{Method}

In this section, we elaborate two implementations for mitigating the trade-off between standard error and robustness error. The overall structure of our adversarial training framework can be found in Figure \ref{fig:subfig1}. The first one is in-line with other adversarial training methods such as PGD-AT \cite{madry2017towards}, TRADES \cite{zhang2019theoretically}, where we attempted to generate null space projected samples to train model parameters $\theta^{adv}$ without affecting generalization performance. The second method is to train the last linear layer, ${W}_L^T$, by projecting gradient to the null space, which essentially keep track of the output of $f(x;\theta^{adv})$ and $f(x; \theta^{std})$.  

We have demonstrated a toy sample representation difference between typical adversarial training and null projection-based adversarial training in Figure \ref{ToyDistribution}. For typical adversarial training, adversarial perturbation is unconstrained, resulting in sample crossing decision boundary. Whereas for null projection-based adversarial training, the perturbation is constrained to $Null({W^{std}})$ which is orthogonal to the space affecting decision boundary.

\subsection{Null-space Projected Data Augmentation}
Recall Eq.\eqref{NSAT robost training loss general form}, we can rewrite the objective function in this case as,
\begin{equation}
  \begin{aligned}
  \label{NPDA loss}
    \mathcal{\hat{L}}^{robust} &= \min_{{\theta}^{adv}}\max_{\delta}l(f(x+\delta;{\theta}^{adv}), y)\\
    where \quad & f(x;\theta^{std}) = f(x+\delta;\theta^{std})
\end{aligned}
  \end{equation}

That is, we intend to search for $\delta$ that keep the model output identical, while minimizes boundary error as much as possible. From Eq.\eqref{std model expansion} and Eq.\eqref{adv model expansion}, if ${W^{std}}_L^T \Delta h^{adv}=0$, constraint term in Eq.\eqref{NPDA loss} holds.
Recall the definition of null space, it means that $\Delta h^{adv}$ maps to null space of ${W^{std}}_L^T$, $Null({W^{std}}_L^T)$. We will abbreviate it as  $Null({W^{std}})$ in the following sections.

Equivalently, we can represent Eq.\eqref{NPDA loss} as,
\begin{equation}
  \begin{aligned}
  \label{NPDA loss alternative form}
    \mathcal{\hat{L}}^{robust} &= \min_{{\theta}^{adv}}\max_{\delta \rightarrow \Delta h^{adv} \in Null({W^{std}})}l(f(x + \delta;{\theta}^{adv}), y)\\
\end{aligned}
  \end{equation}
The disturbance incorporates precise parameter gradient information from the current training model, thereby augmenting the model's robustness against adversarial attacks relying on reverse gradients. Furthermore, this perturbation is carefully restricted within the null-space of a well-established model, ensuring that it does not have a negative repercussion on the optimal accuracy for non-disturbed samples.


 However, it is tough to directly find a $\delta$, which maps to $\Delta h^{adv} \in Null({W^{std}})$. Alternatively, we can generate it reversely. Firstly, we can generate derivatives with respect to $h_{L-1}$ and project it to $Null({W^{std}})$ to form a null projected adversarial representation in penultimate layer $h_{L-1}$.
\begin{equation}
  \begin{aligned}
  \label{pertubation generation in penultimate layer}
 \mathcal{T}_h^{adv-np} : h_{L-1}^{adv-np}&=h^{std}+\eta P_{Null({W^{std}})} \cdot \frac{\partial l}{\partial y} \cdot \frac{\partial y}{\partial h_{L-1}}\\
                                &=h^{std}+\eta P_{Null({W^{std}})} \cdot \frac{\partial l}{\partial y}\cdot W_L^T  
\end{aligned}
  \end{equation}
Having generated $h_{L-1}^{adv-np}$, we can compute “equivalent” adversarial sample by carrying on applying chain-rule.
\begin{algorithm}[tb]
\caption{Adversarial Training by Null Projected Data Augumentation}
\label{alg:algorithm}
\textbf{Input:}Step sizes $\eta_{1}$ and $\eta_{2}$, batch size $m$, number of iteration K in inner optimization, network architecture parameterized by $\theta^{adv}$\\
\textbf{Output:} Robust network $f(\cdot; \theta^{adv})$\\
\begin{algorithmic}[1]
\STATE Initialize network $f(\cdot; \theta^{std})$ with standard training configuration
\STATE W = GetLinearWeight($f(\cdot;{\theta}^{std})$)
\STATE $P_{N}^{W}$ = ComputeNullProjectionMatrix(W)
\REPEAT
\STATE Read mini-batch $B = \{x_{1},...,x_{m}\}$ from training set
\FOR{$i=1,...,m (in parallel)$}
    \STATE    $x_{i}' \leftarrow x_{i} + 0.001 \cdot \mathcal N(\bf 0, \bf I)$, where $N(\bf 0, \bf I) $ is the Gaussian distribution with zero mean and identity variance
        \FOR {$k=1,...,K$}
            \STATE $\ell = \mathcal L (f_{\theta}(x_{i}), y)$
            \STATE $\Delta x = (P_{N}^{W} \cdot (\frac{\partial \ell}{\partial h})^{\text{T}})^{\text{T}} \cdot \frac{\partial h}{\partial x}$, where $h$ is the last hidden layer before mapping to $y = softmax(W^{\text{T}}\cdot h)$
            \STATE $x_{i}' \leftarrow  \prod_{B(x_{i}, \epsilon)} x_{i}' + \eta_{1} \cdot sign(\Delta x)$
        \ENDFOR
    \ENDFOR 
    \STATE $\theta^{adv} \leftarrow \theta^{adv} - \eta_{2} \sum_{i=1}^{m}\nabla_{\theta^{adv}} \mathcal L (f(x_{i}), y)$ 
\UNTIL {training converged}
\end{algorithmic}
\end{algorithm}

\begin{equation}
  \begin{aligned}
  \label{pertubation generation for input}
\mathcal{T}_x^{adv-np}:x^{adv-np}&= x^{std}+ \eta \frac{\partial l}{\partial y} \cdot \frac{\partial y}{\partial h_{L-1}}\cdot \frac{\partial h_{L-1}}{\partial x}\\
&= x^{std}+ \eta P_{Null({W^{std}})} \frac{\partial l}{\partial y} W_L^T  \frac{\partial h_{L-1}}{\partial x}
\end{aligned}
  \end{equation}
The adversarial sample is then generated as \cite{madry2017towards} iteratively,
\begin{equation}
  \begin{aligned}
  \label{iterative perturbation generation}
x_{t+1}^{adv-np}= \prod_{B(x,\varepsilon)} x_t^{adv-np}+\eta P_{Null({W^{std}})} \frac{\partial l}{\partial y} W_L^T \frac{{\partial h}_{L-1}}{\partial x}
\end{aligned}
  \end{equation}

\begin{algorithm}[t]
\caption{Adversarial Training by Null Projected Gradient Descent}
\label{algorithm}
\textbf{Input:} Step sizes $\eta_{1}$ and $\eta_{2}$, batch size $m$, number of iteration K in inner optimization, network architecture parameterized by $\theta^{adv}$, number of layer L in the network architecture \\
\textbf{Output:} Robust network $f(\cdot;\theta^{adv})$
\begin{algorithmic}[1]
\STATE Initialize network $f(\cdot;\theta^{std})$ with standard training configuration
\STATE W = GetLinearWeight($f(\cdot;\theta^{std})$)
\STATE $P_{N}^{W}$ = ComputeNullProjectionMatrix(W)\;
\REPEAT 
    \STATE Read mini-batch $B = \{x_{1},...,x_{m}\}$ from training set
    \FOR{$i=1,...,m (in parallel)$}
    
        \STATE $x_{i}' \leftarrow x_{i} + 0.001 \cdot \mathcal N(\bf 0, \bf I)$, where $N(\bf 0, \bf I) $ is the Gaussian distribution with zero mean and identity variance
        \FOR {$k=1,...,K$}
        
            \STATE $x_{i}' \leftarrow  \prod_{B(x_{i}, \epsilon)} x_{i}' + \eta_1 \cdot sign(\nabla_{\theta} \mathcal L (f(x_{i}), y))$
        \ENDFOR
    \ENDFOR
    \STATE $\ell = \mathcal L (f_{\theta}(x_{i}), y)$\;
    \FOR{j=L,...,1}
        \IF{n = L}
            \STATE $ W^{n} = W^{n} - \eta_{2} \cdot (P_{N}^{W} \cdot \frac{\partial \ell}{\partial W})$
        \ELSE
            \STATE $ W^{n} = W^{n} - \eta_{2} \cdot ( \frac{\partial \ell}{\partial h} \cdot \frac{\partial h}{\partial W^{n}})$, where $h$ is the last hidden layer before mapping to $y = softmax(W^{\text{T}}\cdot h)$
        \ENDIF
    \ENDFOR
\UNTIL {training converged}
\end{algorithmic}
\end{algorithm}

In Figure \ref{fig:subfig2}, we have shown a multi-step null projected sample generation process climbing up the hill of loss landscape. The gradient updated of NPDA in each layer is in Appendix C. In algorithm 1, we illustrate the detailed steps of NPDA. In \textbf{step 2\&3}, we generate $P_{Null({W^{std}})}$ by the weight of last linear layer of the pretrained high-accurate model $f(\cdot;\theta^{std})$. We generate a batch of training samples (\textbf{step 5}) with small Guassian noise (\textbf{step 7}). The adversarial loss is computed in a feed-forward prediction (\textbf{step 9}) and the perturbation noise is computed as Eq.\eqref{pertubation generation for input} in \textbf{step 10} and added to get adversarial sample in \textbf{step 11}. The parameter is updated as usual by adversarial loss in \textbf{step 14}. Notice that the loss is replaceable for any existing adversarial loss.

\subsection{Null-space Projected Gradient Descent}
Again, we initiate from our objective function in general form in Eq.\eqref{NSAT robost training loss general form}. This time instead of imposing constraint on $\delta$, we cast constraint on model parameter $\theta$. Since we start the adversarial training with the standard training model, $f(\cdot;{\theta}^{adv})|_{t=0} = f(\cdot;{\theta}^{std})$, where $t$ denotes the number of epochs trained. We can relax the the objective function in this scenario as,
\begin{equation}
  \begin{aligned}
  \label{NPGD loss}
    \mathcal{\hat{L}}^{robust} &= \min_{{\theta}^{adv}}\max_{\delta}l(f(x + \delta; {\theta}^{adv}), y)\\
    s.t. \quad & f(x;{\theta}^{std}) \approx f(x;{\theta}^{adv})
\end{aligned}
  \end{equation}

Likewise, the gradient update in the non-linear layers are trivial for us, as we only interested in the last layer to keep track of the constraint term in Eq. \eqref{NPGD loss}. In this way, we train ${W^{adv}}_L^T$ simply by projecting the derivative to the null space, $Null({W^{std}}^T_L)$. The remaining settings are implemented as standard adversarial training. 

\begin{remark}
    \label{remark 3.2}
    The error between adversarial training model $f(x;\theta^{adv})$ trained by NPGD and $ f(x;\theta^{std})$ is an element belongs to the null space of $W^{std}$, $Null({W^{std}})$.
    \begin{equation}
  \begin{aligned}
        f(x;\theta^{adv}) - f(x;\theta^{std}) \in Null({W^{std}})
  \end{aligned}
  \end{equation}
\end{remark}

See Appendix.B.1 for detailed proof of Remark \ref{remark 3.2}.\\
Analogously, we compute the null projection matrix (\textbf{step 3}) from last linear layer of pretrained high-accurate pretrained model (\textbf{step 2}) and subsequently generate a batch of training samples (\textbf{step 5}) with small Guassian noise (\textbf{step 7}). The adversarial samples are generated iteratively towards gradient ascent direction (\textbf{step 10}). Again, the loss for generating adversarial samples are replaceable to any State-Of-The-Art adversarial loss in the literature. The gradient updated for ${W^{adv}}_L^T$ can be represented as Eq.\eqref{linear layer gradient update} in \textbf{step 15}.

\begin{equation}
  \begin{aligned}
  \label{linear layer gradient update}
  {W^{adv}}_L^T  &\leftarrow {W^{adv}}_L^T-\eta P_{Null({W^{std}})} \cdot \frac{\partial l}{\partial y}\cdot \frac{\partial l}{\partial {W^{adv}}_L^T}\\ 
  &={W^{adv}}_L^T-\eta P_{Null({W^{std}})} \cdot \frac{\partial l}{\partial y}\cdot h_{L-1}^{adv}  
\end{aligned}
  \end{equation}
  
Gradient updated for a particular layer $W_n$ (\textbf{step 17}) is equivalent to standard adversarial training method. However, it does not mean the gradient updated is identical to that of standard adversarial training. The gradient for a particular layer $W_n$ follows the change of $W_L^T$ in the following steps.

\begin{equation}
  \begin{aligned}
  \label{non-linear layer gradient update}
W_n^T  \leftarrow W_n^T-\eta \cdot \frac{\partial l}{\partial h}\cdot \frac{\partial h}{\partial W_n}
\end{aligned}
  \end{equation}

\begin{table*}[t]
 \centering
\begin{adjustbox}{width=2\columnwidth,center}
    \begin{tabular}{llllrrrrrr}
    \toprule
    Dataset						&						&			& 					& \multicolumn{3}{c}{CIFAR10} 				  & \multicolumn{3}{c}{SVHN} \\
								&	Adv Gen. Method		& 	Loss	& Pretrained Model 	&	Clean Error & PGD Error & AA Error & Clean Error & PGD Error & AA Error  \\
    \midrule
    st model 					&						&	CE		& ImageNet 			& 5.69\% 		& 84.28\% 			& 84.18\% & 4.43\% & 98.01\% & 93.77\% \\
    lse st model 				&						&	LSE		& ImageNet 			& 5.66\% 		& 84.81\% 			& 84.29\% & 4.34\% & 96.32\% & 93.51\% \\
    PGD-AT 						&	PGD					&	PGD		& st model 			& 12.38\% 		& 23.64\% 			& 23.72\% & 6.21\% & 26.70\% & 27.30\% \\
    TRADES 						&	TRADES				&	TRADES	& st model 			& 12.01\% 		& 26.59\% 			& 26.73\% & 6.93\% & 39.74\% & 40.58\% \\
    TRADES \(@\beta=0.01\)			&	TRADES				&	TRADES	& st model 			& 6.95\% 		& 62.13\% 			& 63.70\% & 4.12\% & 87.04\% & 90.55\% \\
    FRL + Reweight + Remargin 	&	TRADES				& 	FRL	    & st model 			& 12.56\% 		& 26.96\% 			& 27.08\% & 6.63\% & 38.82\% & 39.67\% \\
    CAAT  						&	CAAT				&	CAAT	& st model 			& 11.66\% 		& 22.00\% 			& 22.05\% & 5.96\%  & 30.27\% & 30.72\% \\
    SCORE 						&	SCORE				&	SCORE	& st model 			& 12.78\% 		& 34.02\% 			& 34.42\% & 6.68\%  & 43.74\% & 44.95\% \\
    \midrule
    NPDA + PGD-AT  				&	PGD					&	TRADES	& st model 			& 6.58\% 		& 39.37\% 			& 39.40\% & 4.43\% & 79.86\% & 79.79\% \\
    NPDA + TRADES 				&	TRADES				&	TRADES	& st model 			& 6.38\% 		& 43.27\% 			& 43.27\% & 4.89\% & 84.04\% & 84.02\% \\
    NPDA + SCORE 				&	SCORE				&	TRADES	& st model 			& 5.96\% 		& 46.46\% 			& 46.48\% & 4.94\% & 89.67\% & 89.70\% \\
    \midrule
    NPGD + PGD-AT 				&	PGD					&	TRADES	& st model 			& 7.04\% 		& 26.41\% 			& 28.01\% & 4.05\% & 34.97\% & 36.87\% \\
    NPGD + TRADES  				&	TRADES				&	TRADES	& st model 			& 6.54\% 		& 28.17\% 			& 28.37\% & 4.05\% & 41.59\% & 42.61\% \\
    NPGD + SCORE  				&	SCORE				&	TRADES	& lse st model 		& 6.60\% 		& 34.52\% 			& 34.89\% & 4.06\% & 44.75\% & 46.19\% \\
    NPGD + SCORE  				&	SCORE				&	TRADES	& st model 			& 6.25\% 		& 35.85\% 			& 36.18\% & 4.19\% & 44.00\% & 45.42\% \\
    \bottomrule
    \end{tabular}%
  \end{adjustbox}
  
\caption{Comparison of Standard Error \& Robustness Error for Models on CIFAR10 \& SVHN.}
\label{tab:table1}%
\end{table*}%

\section{Experiments \& Evaluation}
\textbf{Experiment Setups:} We have adopted CIFAR10 and SVHN to verify the effectiveness of our methods. The backbone model used in the experiment were kept with Pre-Act Resnet in this work. During Training, the adversarial samples were found by iterating 10 steps and the adversarial attack coefficient $\eta$ in each step and pre-defined adversarial bound were $2/255$ and $10/255$. We have implemented PGD attack and Auto-attack \cite{croce2020reliable} for testing robustness. The adversarial bound in the test phase for both PGD attack and Auto-attack is $8/255$. Notice that we have normalized input data with mean and standard deviation. We set mean to $125.3, 123.0, 113.9$ and standard deviation to $63.0, 62.1, 66.7$ for CIFAR10 and set both mean and standard deviation to $0.5$ for SVHN, which is different from the setting as \cite{croce2020robustbench} but followed the same experiment setup to \cite{zhou2023combining} for comparison. All our experiments were implemented on a NVidia V100 GPU.\\
We have compared State-Of-The-Art (SOTA) adversarial training methods such as PGD-AT, TRADES, FRL, SCORE, CAAT. \cite{zhang2019theoretically}\cite{xu2021robust}\cite{madry2017towards}\cite{zhou2023combining} Since our methods can be utilized seamlessly with loss defined by these SOTA methods, we report our results accordingly. PGD was used as adversarial method in all our testing.
\subsection{Main Results}

\begin{figure}[h]
\subfloat[Training Dynamics of Loss]{
    \label{fig6:a}\includegraphics[width=0.5\columnwidth]{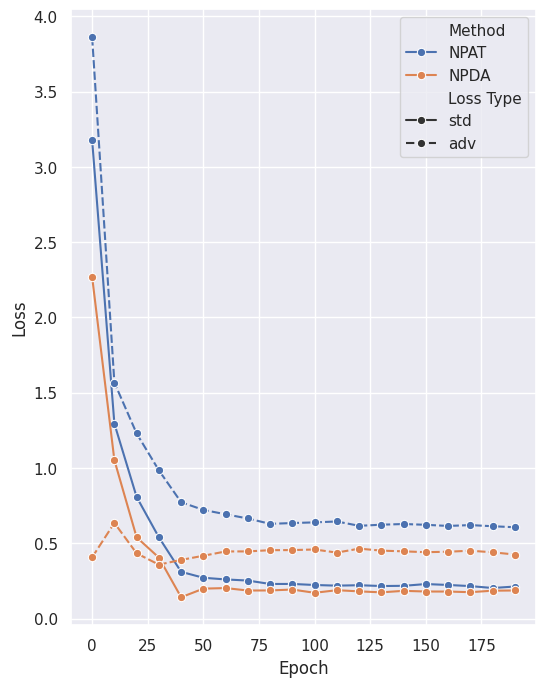}
}
\subfloat[Training Dynamics of Accuracy \& Robustness ]{
    \label{fig6:b}\includegraphics[width=0.5\columnwidth]{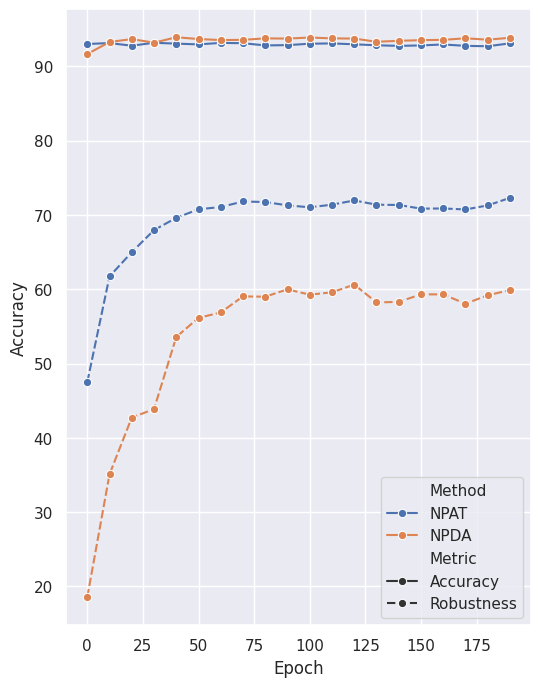}
}
\caption{Loss and Accuracy \& Robustness Training Dynamics for 200 Epochs}
\label{figure6}
\end{figure}

We first report the overall performance of different models evaluated on CIFAR10 and SVHN datasets in Table \ref{tab:table1} and a scatter plot of performance of CIFAR10 is shown in Figure \ref{ScatterPlot}. The st model and lse st model are two standard training models trained by Cross-Entropy(CE) loss and least-squared error(LSE) with pretrained parameters on ImageNet \cite{ILSVRC15}. The LSE loss is used in the SCORE method as the classification loss. As TRADES loss splitted the CE loss to a classification loss and boundary loss, it allows us imposing an adversarial coefficient $\beta$ to control the level of trade-off between generalization and robustness, whereas the extent of robustness is arbitrary for CE loss. Thus, for a fair comparison, we used TRADES loss for the baseline adversarial models in most cases except for PGD, CAAT and SCORE. 
\\
To validate the effectiveness of our model, we have tuned $\beta$ to $0.01$ to obtain similar clean error as NPDA and NPGD. Both our null-space projector based methods outperform that of baseline TRADES@$\beta=0.01$, showing that our boost in robustness is indeed not a result of controlling the level of adversarial loss by tuning hyper-parameter $\beta$.\\
The standard error of NPDA \& NPGD under most of configurations are close to that of standard training, except for the lse st model parameter initialized NPGD with SCORE loss and outperform all adversarial baseline methods. The maximum difference between our null-space projected method and standard model are 1.35\% and 0.6\% for CIFAR10 \& SVHN. In general, we have observed a minor accuracy drop except for NPGD on SVHN. Neither NPDA nor NPGD outperforms each other consistently in both datasets in terms of accuracy. \\
Without losing too much on standard accuracy, NPGD obtained a comparable robustness error. The best robustness errors among all adversarial baselines are from PGD and our method NPGD reached almost the same level for CIFAR10 comparing with the best case of baseline adversarial methods, whereas there was a 35.83\% gap for SVHN without hindering the generalization performance. \\
We illustrated the training dynamics for NPDA and NPGD in Figure \ref{figure6}. There is no trade-off between generalization and robustness in terms of losses and accuracy, which evidently show that we obtained extra robustness without sacrificing generalization under the scope of no extra dataset and optimizing model structure. 


\begin{table*}[t]
 \centering
\begin{adjustbox}{width=2\columnwidth,center}
    \begin{tabular}{llrrrrrr}
    \toprule
	Dataset 					&   				& \multicolumn{3}{c}{CIFAR10} & \multicolumn{3}{c}{SVHN} \\
								& Hidden Size   	& Clean Error & PGD Error & AA Error & Clean Error & Clean PGD Error & AA Error  \\
    \midrule
	\multirow{4}{*}{NPDA}		&	512				&	6.58\%	&39.37\%	&39.86\%	&4.43\%	&79.86\%	&79.79\%\\
								&	1024			&	6.11\%	&39.87\%	&55.73\%	&3.75\%	&71.07\%	&71.09\%\\
								&	2048			&	7.39\%	&48.34\%	&56.84\%	&3.79\%	&55.08\%	&55.04\%\\
								&	4096			&	6.96\%	&35.24\%	&42.72\%	&3.90\%	&55.79\%	&55.74\%\\

	\midrule
	\multirow{4}{*}{NPGD}	&	512				&	7.04\%	&26.41\%	&28.01\%	&4.05\% &34.97\%	&36.87\%\\
								&	1024			&	6.67\%	&30.03\%	&30.21\%	&4.24\%	&36.96\%	&38.58\%\\
								&	2048			&	7.59\%	&51.14\%	&51.19\%	&4.20\%	&28.90\%	&30.07\%\\
								&	4096			&	6.85\%	&38.10\%	&38.13\%	&4.09\%	&30.59\%	&32.05\%\\

	\bottomrule
    \end{tabular}%
  \end{adjustbox}
\caption{Variation of Hidden Size}
\label{table3}%
\end{table*}%

\subsection{Variation of Adversarial Coefficient $\beta$}
\begin{figure}[h]
\subfloat[CIFAR10]{
    \label{fig4:a}\includegraphics[width=0.5\columnwidth]{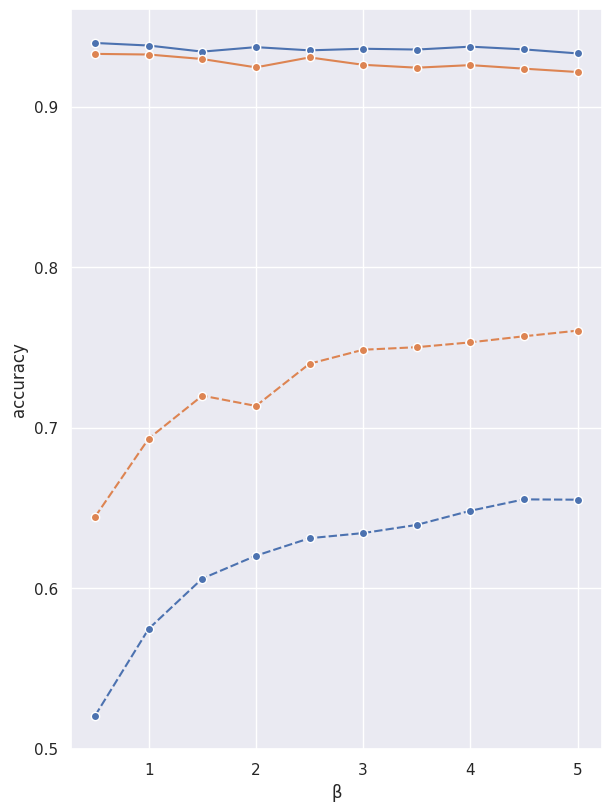}
}
\subfloat[SVHN]{
    \label{fig4:b}\includegraphics[width=0.5\columnwidth]{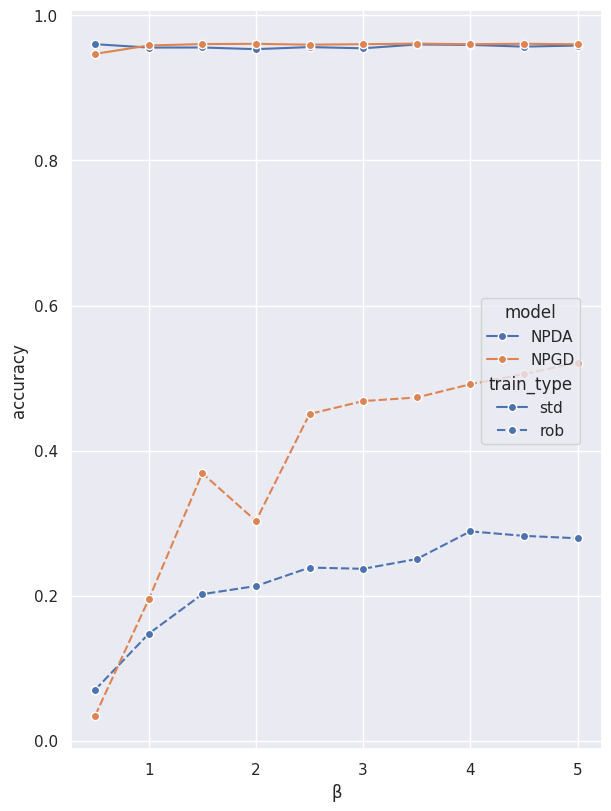}
}
\caption{Variation of Accuracy \& Auto-attack Robustness w.r.t Adversarial Coefficient $\beta$. }
\label{figure4}
\end{figure}

\begin{figure}[t]
\subfloat[Adversarial Attack]{
    \label{fig5:a}\includegraphics[width=0.5\columnwidth]{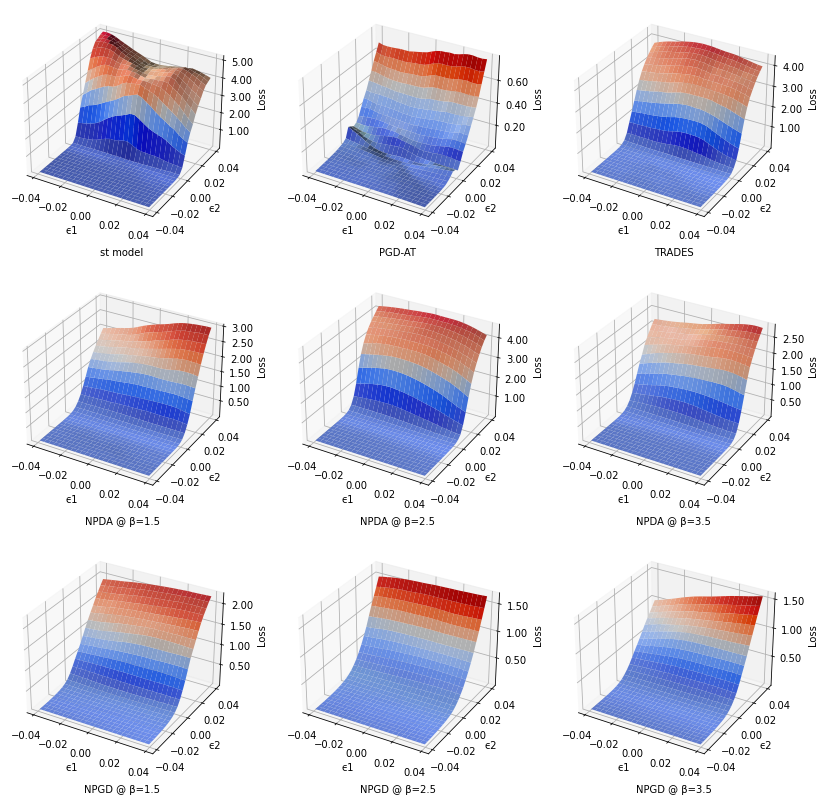}
}
\subfloat[Random Attack]{
    \label{fig5:b}\includegraphics[width=0.5\columnwidth]{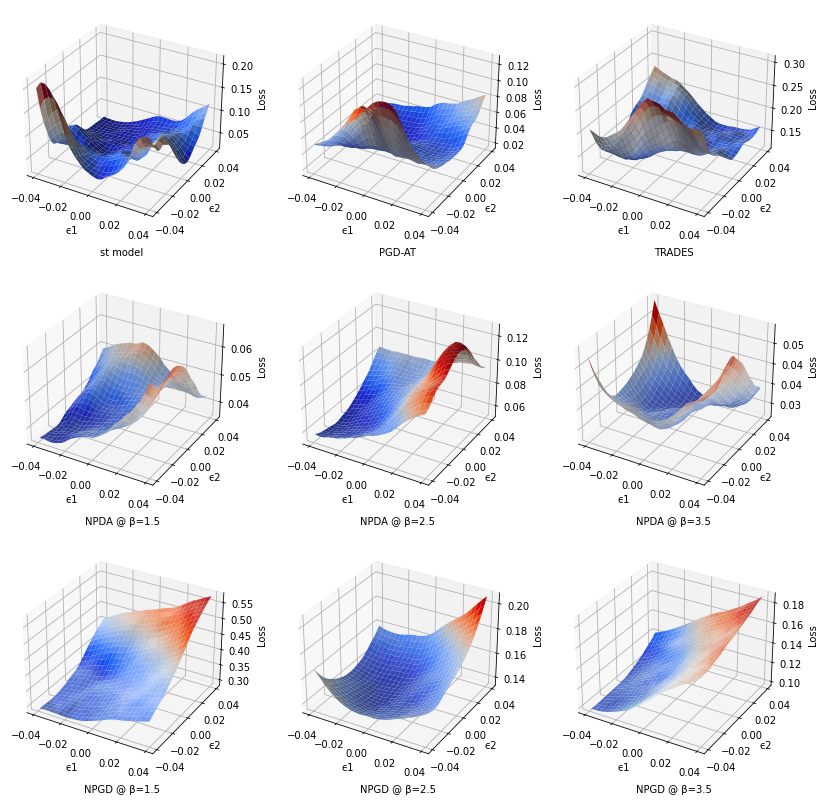}
}
\caption{Loss Landscape of Different Models. st model stands for the model trained under standard training. The landscape of adversarial attack is plotted with one direction of gradient and one random direction. The landscape of random attack is plotted with two random direction.}
\label{figure5}
\end{figure}

We then experimented on variation of different adversarial coefficient $\beta$ to see if it is possible to improve robustness error without hurting standard error. The adversarial sample generation method was PGD-AT and the loss used was TRADES for all cases. In Figure \ref{figure4}, the accuracy on both datasets are almost straight lines with negligible drop, when increasing adversarial coefficient $\beta$. As a result, we still see a trade-off as we gradually increase $\beta$. The cost in trading off robustness for standard error is considerably low under this scope and the robustness gradually saturates as $\beta$ increases. The detailed experimental result can be found in Appendix D.\\
We plotted the loss landscape of PGD-AT, TRADES, NPDA under different adversarial coefficient $\beta$ and NPGD under different adversarial coefficient $\beta$ with adversarial attack and random attack in Figure \ref{figure5}. From our observation, the adversarial training methods generally produce a smoother landscape and by increasing adversarial coefficient, the loss landscape of NPDA \& NPGD become smoother.

\subsection{Variation of Hidden Size}

Lastly, we investigated the size of the null space by changing hidden size of penultimate layer. As we attempted to increase the null space for the same standard trained model, we initialized models with same backbone for all ResNet blocks and introduced an extra linear layer with different hidden sizes in testing. In general, we observed better robustness with larger hidden size for SVHN but the robustness fluctuated on CIFAR10 from Table \ref{table3}, while the generalization on both datasets were around the same level under different hidden sizes.

\section{Related Works}
There are many previous works provided thorough analysis and existence of the accuracy-robustness trade-off problem. \cite{tsipras2018robustness} claimed that the trade-off is inevitable as the objective of two tasks are fundamentally different. They showed the difference by showing a simplified example composing of a moderately correlated robust feature, and a set of strongly correlated vulnerable features altogether grouped as a “meta” feature. The optimum accuracy cannot be reached without utilizing the “meta” feature. \cite{fawzi2018analysis} proposed a framework of analyzing the trade-off for linear classifier and quadratic classifier. On the contrary, \cite{nakkiran2019adversarial} proposed it is the capacity of model that determines the level of robustness. \cite{croce2020robustbench} have proposed a standard robust bench with promising generalization and robustness, but  methods with high rank introduces extra dataset or search for a model capacity by neural architecture search(NAS). 

The earliest adversarial training method, PGD-AT, was proposed by \cite{madry2017towards} and adversarial training was proven to be the most effective way of improving model robustness by \cite{athalye2018obfuscated}. \cite{zhang2019theoretically} designed a trade-off loss, (aka. TRADES) by splitting the standard loss and adversarial loss. However, it is an over-strong assumption that all robust features can be learned by model, which might not be the case in reality due to the model architecture and the way of training. \cite{raghunathan2020understanding} also used a noiseless linear regressor to show effect of parameter error when introducing extra dataset (adversarial samples). They provided three theoretical conditions to avoid the trading-off and proved the effectiveness of Robust Self Training (RST) method. Nonetheless, their conditions are for linear model and difficult to meet for generating adversarial samples. \cite{pang2022robustness} declared the trade-off is partially due to the misalignment of learned adversarial estimator $p_{\theta^\ast}(y|x)$ and joint data distribution $p_d(y|x)$, and proposed a Self-Consistent Robust Error (SCORE) loss by reformulating adversarial loss.

The other source of error is known as unfairness, as there exists disparity of samples among different classes due to unequal variance, priors and noise level. \cite{xu2021robust} attempted to leverage the fairness by continuously estimating the upper bound of boundary error and reweighting sample loss for each class (FRL). Essentially, it forms unequal decision boundaries between classes. Upon FRL, \cite{zhou2023combining} introduced an anti-adversarial sample-based method, CAAT, to cope with issue of noisy-sample. 
The adversarial training task can be considered as a multi-task learning problem since the extra adversarial samples are under same distribution as original dataset. There is a notorious catastrophic forgetting problem where model performance degrades on previous tasks when learning on new task. \cite{kirkpatrick2017overcoming} proposed a regularizer-based method, EWC, penalizing large deviation parameters from previous tasks. \cite{wang2021training} have proposed a null-space projecting optimizer for continual learning, which performs null space estimation based on space of previous parameters and the null projection were deployed to every layer of the model. 

\cite{ravfogel2020null} proposed an iterative null-projection method for removing sensitive information from the representation and obtaining an exclusive estimator. Our work is greatly inspired by the way of decomposing model in their work.

\section{Conclusion}
In this work, we provided theoretical studies of training an adversarial estimator in terms of its non-linear backbone and last linear transformation. We then proposed two methods accordingly with derivation of gradient update in both cases. Finally, we verified our methods under different settings to reveal the effectiveness on CIFAR10 and SVHN datasets.

\newpage
\bibliography{main}
\bibliographystyle{icml2024}

\newpage
\appendix
\onecolumn

\section*{Appendix}

\section{Null-Space Projector By SVD}
Suppose we have a matrix $W\in R^{m\times n}$ and $r(W) < \min(m, n)$ factorized by SVD, $W=U\Sigma V^T \in R^{m \times n}$, where $U\in R^{m\times m}$ corresponds to orthonormal basis of the column space of $W$, $\Sigma \in R^{m \times n}$ is a pseudo-diagonal matrix. $V^T \in R^{n\times n}$ is the orthonormal basis of row space of $W$.

Recall Definition 2.1 \& Definition 2.2, there exists a $P_{Null_(W)}$ that satisfies $WP_{Null(W)}x = 0, \quad for \quad \forall x \in R^{n \times 1}$.

By factorizing $W$ and substitute closed form solution of $P_{Null(W)}$ from SVD based on Eq.\eqref{SVD null space solution}, we have,
\begin{equation}
  \begin{aligned}
    WP_{Null(W)}x &= U\Sigma V^T(I-VV^T)\cdot x\\
                  &= (U\Sigma V^T - U\Sigma V^TVV^T)\cdot x, \text{where} \quad V^TV = I\\ 
                  &= (U\Sigma V^T - U\Sigma V^T)\cdot x\\
                  &= 0\cdot x\\
                  &= 0
\end{aligned}
  \end{equation}

\section{Theoretical Guarantee of NSAT}

\subsection{Proof of Remark 3.1}
\begin{proof}
From Eq. \eqref{linear layer gradient update}, we are updating the parameter of last linear layer, $W^T_L$ in a mini-batch as,
\begin{equation*}
\begin{split}
    \hat{W}^{s} & = \hat{W}^{s-1} + \epsilon P_{Null(W)} \widehat{\mathrm{item}}(s-1) \\
                & = \hat{W}^{s-2} + \epsilon P_{Null(W)} (\widehat{\mathrm{item}}(s-1) + \widehat{\mathrm{item}}(s-2)) \\
                & = W^{std} + \epsilon P_{Null(W)} \sum_{k=0}^{s-1} \widehat{\mathrm{item}}(k).
\end{split}
\end{equation*}
where $\widehat{\mathrm{item}}$ are partial derivatives computed from each batch of data.\\
Eventually, we get optimal $W^T_L$ as,
\begin{equation}
    \hat{W}^{\mathrm{opt}} = W^{std} + \epsilon P_{Null(W)} \sum_{i}^{stop} \widehat{\mathrm{item}}(i),
\end{equation}
where the last summation term are mapped to the null space of W, $\mathrm{Null}(W)$.
Since the $\hat{W}^{\mathrm{opt}}$ is the last linear layer of $f_\theta^{adv}(x)$ trained by NPGD. Thereby, we have 
\begin{equation}
  \begin{aligned}
  \label{f_NPGD expansion}
   f_\theta^{adv}(x) &= [W^{std} + \epsilon P_{Null(W)} \sum_{i}^{stop} \widehat{\mathrm{item}}(i)]H\\
   &= W^{std}H + \epsilon P_{Null(W)} \sum_{i}^{stop} \widehat{\mathrm{item}}(i) H\\
   &= f_\theta^{std}(x) + \epsilon P_{Null(W)} \sum_{i}^{stop} \widehat{\mathrm{item}}(i) H
  \end{aligned}
  \end{equation}

Rearrange Eq. \eqref{f_NPGD expansion},
\begin{equation}
  \begin{aligned}
f_\theta^{adv}(x) - f_\theta^{std}(x) &= \epsilon P_{Null(W)} \sum_{i}^{stop} \widehat{\mathrm{item}}(i) H \in Null(W)
  \end{aligned}
  \end{equation}
\end{proof}

\section{Gradient Update of NPDA}
The gradients with respect to $W_L^T$ and $H_{L-1}$ are shown in Eq. \eqref{NPDA W gradient} and Eq. \eqref{NPDA H gradient}.
\begin{equation}
  \begin{aligned}
  \label{NPDA W gradient}
\frac{\partial l}{\partial W_L^T}=\frac{\partial l}{\partial y}\cdot \frac{\partial y}{\partial W_L^T} = \frac{\partial l}{\partial y} \cdot {h_{L-1}^{adv}}
\end{aligned}
  \end{equation}

\begin{equation}
  \begin{aligned}
  \label{NPDA H gradient}
\frac{\partial l}{\partial h_{L-1}}=\frac{\partial l}{\partial y} \cdot \frac{\partial y}{\partial h_{L-1}}= \frac{\partial l}{\partial y} \cdot W_L^T
\end{aligned}
  \end{equation}
  
The gradient updated for $W_L^T$ can be represented as in Eq. \eqref{NPDA W gradient update}.
\begin{equation}
  \begin{aligned}
  \label{NPDA W gradient update}
W_L^T  \leftarrow W_L^T- \eta \frac{\partial l}{\partial y} \cdot h_{L-1}^{adv}
\end{aligned}
  \end{equation}

Next, let us elaborate the gradient updated in each layer. The gradient computed for a particular layer $W_n$ for $0<n\le L-1$ can be represented as,

\begin{equation}
  \begin{aligned}
\frac{\partial l}{\partial W_n^T}&=\frac{\partial l}{\partial y}\cdot \frac{\partial y}{\partial h_{L-1}}\cdot \frac{\partial h_{L-1}}{\partial h_n}\cdot \frac{\partial h_n}{\partial W_n^T}\\
&=\frac{\partial l}{\partial y}\cdot W_L^T \cdot \frac{\partial h_{L-1}}{\partial h_n}\cdot \frac{\partial h_n}{\partial W_n^T}\\
&=\frac{\partial l}{\partial y}\cdot W_L^T \cdot \frac{\partial h_{L-1}}{\partial h_n}\cdot h_{n-1}^{adv}
\end{aligned}
  \end{equation}

Therefore, the gradient for a particular layer $W_n$ after null space projection is as illustrated in Eq. \eqref{NPDA non-linear layer gradient update}.
\begin{equation}
  \begin{aligned}
  \label{NPDA non-linear layer gradient update}
\frac{\partial l}{\partial W_n^T}=\frac{\partial l}{\partial y}\cdot W_L^T\cdot \frac{\partial h_{L-1}}{\partial h_n }\cdot h_{n-1}^{adv-np}
\end{aligned}
  \end{equation}

\newpage
\section{Variation of Adversarial Coefficient $\beta$}
\begin{table*}[h]
 \centering
\begin{adjustbox}{width=\columnwidth,center}
    \begin{tabular}{llrrrrrr}
    \toprule
	Dataset & 		  & \multicolumn{3}{c}{CIFAR10} & \multicolumn{3}{c}{SVHN} \\
			& $\beta$ & Test Error & Test Robust Error & AA Error & Test Error & Test Robust Error & AA Error  \\
    \midrule
	\multirow{8}{*}{NPDA}		& 	0.5	&	6.04\%	&	47.90\%	&	47.94\%	&	3.98\%	&	92.99\%	&	92.98\%	\\
								& 	1	&	6.20\%	&	42.50\%	&	42.53\%	&	4.44\%	&	85.13\%	&	85.25\%	\\
								&	1.5	&	6.58\%	&	39.37\%	&	39.40\%	&	4.43\%	&	79.86\%	&	79.79\%	\\
								&	2	&	6.30\%	&	38.00\%	&	37.98\%	&	4.66\%	&	78.76\%	&	78.67\%	\\
								& 	2.5	&	6.50\%	&	36.86\%	&	36.89\%	&	4.38\%	&	76.24\%	&	76.13\%	\\
								& 	3	&	6.40\%	&	36.59\%	&	36.57\%	&	4.54\%	&	76.30\%	&	76.30\%	\\
								& 	3.5	&	6.45\%	&	36.05\%	&	36.06\%	&	4.01\%	&	75.01\%	&	74.95\%	\\
								& 	4	&	6.27\%	&	35.19\%	&	35.18\%	&	4.07\%	&	71.12\%	&	71.13\%	\\
								&	4.5	&	6.44\%	&	34.45\%	&	34.47\%	&	4.31\%	&	71.82\%	&	71.76\%	\\
								&	5	&	6.69\%	&	34.48\%	&	34.49\%	&	4.16\%	&	72.14\%	&	72.08\%	\\
	\midrule
	\multirow{8}{*}{NPGD}		&	0.5	&	6.72\%	&	35.17\%	&	35.55\%	&	5.31\%	&	94.26\%	&	96.54\%	\\
								&	1	&	6.76\%	&	30.36\%	&	30.68\%	&	4.16\%	&	73.26\%	&	80.51\%	\\
								&	1.5	&	7.04\%	&	26.41\%	&	28.01\%	&	3.96\%	&	56.21\%	&	63.13\%	\\
								&	2	&	7.56\%	&	28.48\%	&	28.65\%	&	3.93\%	&	63.63\%	&	69.70\%	\\
								&	2.5	&	6.94\%	&	26.04\%	&	26.02\%	&	4.04\%	&	54.84\%	&	54.96\%	\\
								&	3	&	7.40\%	&	25.14\%	&	25.15\%	&	3.98\%	&	53.17\%	&	53.17\%	\\
								&	3.5	&	7.58\%	&	25.01\%	&	24.99\%	&	3.90\%	&	52.64\%	&	52.68\%	\\
								&	4	&	7.42\%	&	24.68\%	&	24.69\%	&	3.98\%	&	50.83\%	&	50.85\%	\\
								&	4.5	&	7.64\%	&	24.32\%	&	24.31\%	&	3.92\%	&	49.50\%	&	49.46\%	\\
								&	5	&	7.85\%	&	23.93\%	&	23.96\%	&	3.99\%	&	47.89\%	&	47.87\%	\\
	\bottomrule
    \end{tabular}%
  \end{adjustbox}
\caption{Variation of Adversarial Coefficient $\beta$}
\label{table4}%
\end{table*}%

\end{document}